\documentclass[preprint,12pt]{elsarticle}

\usepackage{amssymb}
\usepackage[latin1]{inputenc}
\usepackage{amsmath}
\usepackage{multirow}
\usepackage{subfigure}

 \biboptions{comma,square,sort}

\begin{document}

\begin{frontmatter}



 \textbf{NOTICE: this is the author's version of a work that was accepted for publication in \textit{Fuzzy Sets and Systems}. Changes resulting from the publishing process, such as peer review, editing, corrections, structural formatting, and other quality control mechanisms may not be reflected in this document. Changes may have been made to this work since it was submitted for publication. A definitive version was subsequently published in ``Pereira-Fari{\~n}a, M., D\'iaz-Hermida, F., Bugar\'in, A. (2013). On the analysis of set-based fuzzy quantified reasoning using classical syllogistics. \textit{Fuzzy Sets and Systems}, vol. 214(1), 83-94. DOI: 10.1016/j.fss.2012.03.015''}

\title{On the analysis of set-based fuzzy quantified reasoning using classical syllogistics}



\author{M. Pereira-Fariña}
\ead{martin.pereira@usc.es}

\author{F. Díaz-Hermida}
\ead{felix.diaz@usc.es}

\author{A. Bugarín}
\ead{alberto.bugarin.diz@usc.es}

\address{Centro de Investigación en Tecnoloxías da Información (CITIUS), University of Santiago de Compostela, Campus Vida, E-15782, Santiago de Compostela, Spain}

\begin{abstract}
Syllogism is a type of deductive reasoning involving quantified statements. The syllogistic reasoning scheme in the classical Aristotelian framework involves three crisp term sets and four linguistic quantifiers, for which the main support is the linguistic properties of the quantifiers. A number of fuzzy approaches for defining an approximate syllogism have been proposed for which the main support is cardinality calculus. In this paper we analyze fuzzy syllogistic models previously described by Zadeh and Dubois et al. and compare their behavior with that of the classical Aristotelian framework to check which of the 24 classical valid syllogistic reasoning patterns (called moods) are particular crisp cases of these fuzzy approaches. This allows us to assess to what extent these approaches can be considered as either plausible extensions of the classical crisp syllogism or a basis for a general approach to the problem of approximate syllogism.
\end{abstract}



\end{frontmatter}


\section{Introduction}
\label{Sc:Introduction}
Syllogistic inference or syllogism is a type of deductive reasoning in which all the statements involved are quantified propositions. A well-known example of a syllogism is the one shown in Table~\ref{ex:Aristoteliansyllogism}, where $P_{1}$ denotes the major or first premise, $P_{2}$ the minor or second premise, and $C$ the conclusion.

\begin{table}[htb]%
\centering
\begin{tabular} [c]{cc}%
$\left(  P_{1}\right) $ & All human beings are mortal\\
$\left(  P_{2}\right)  $ & All Greeks are human beings\\\hline
$\left(  C\right) $ & All Greeks are mortal
\end{tabular}
\caption{\label{ex:Aristoteliansyllogism} Aristotelian syllogistic inference.}
\end{table}

A quantified proposition involves two main elements: a quantifier (such as \textit{all, many, 25, 25\%, double\dots that of\dots}) and terms, usually interpreted as sets. In the typical binary quantified statement, the subject is the ``restriction'' of the quantifier and the predicate its ``scope''. For instance, for $P_{1}$ in the example in Table~\ref{ex:Aristoteliansyllogism}, \textit{human beings} is the restriction and \textit{mortal} is the scope of the \textit{all} quantifier.\par

\begin{figure}
\centering
\includegraphics[width=0.60\textwidth]{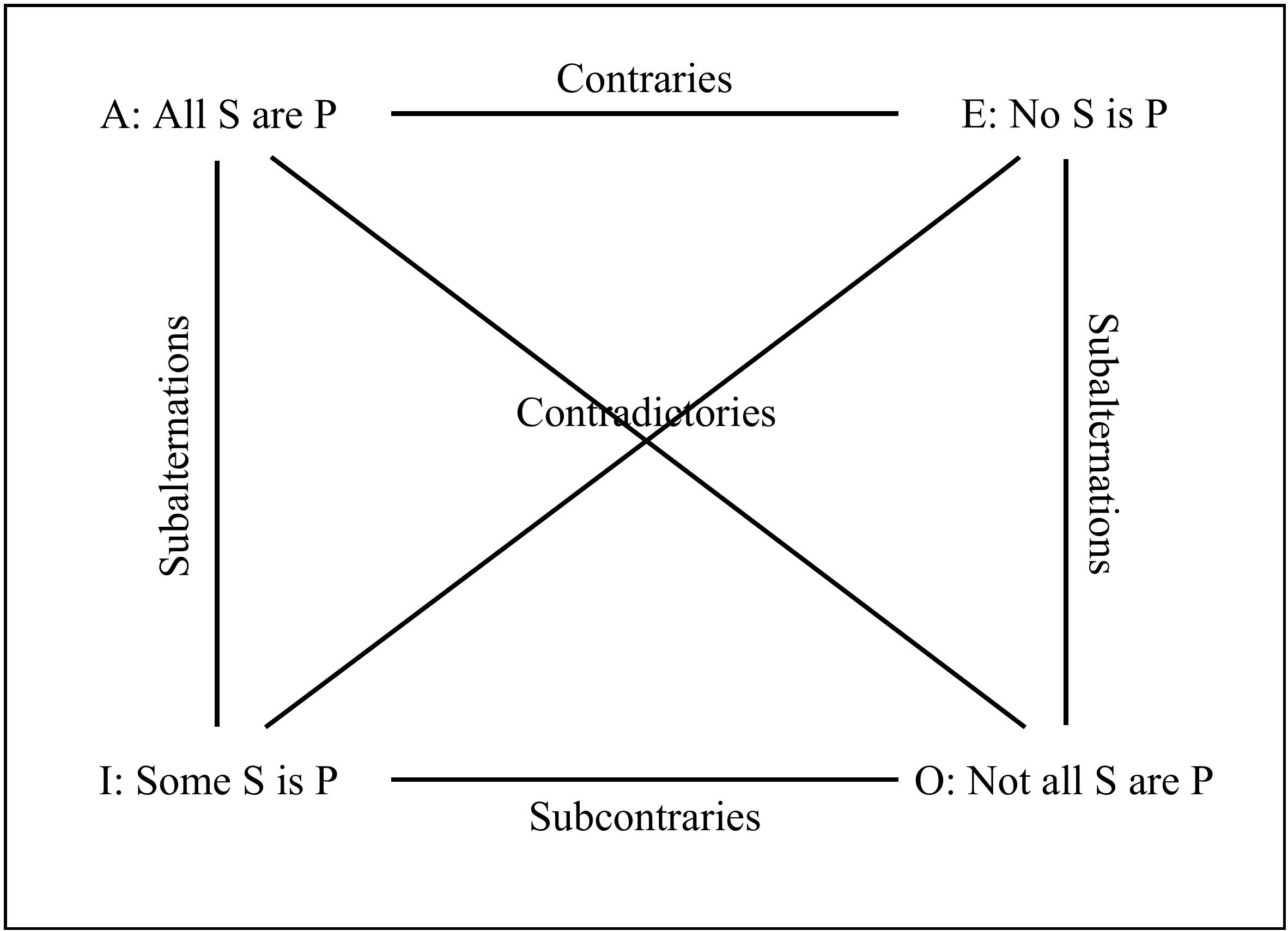}
\caption{\label{fig:LSOclassical} The classical logic square of opposition.}
\end{figure}

The classical approach to syllogism was developed by Aristotle~\cite{Aristotle1949}, who considered four crisp quantifiers: \textit{all} (\textit{A}), \textit{none} (\textit{E}), \textit{some} (\textit{I}), and \textit{not all} (\textit{O}). These quantifiers are defined accordingly to the linguistic properties in the logic square of opposition (LSO)~\cite{Thom1981} shown in Figure~\ref{fig:LSOclassical}.\par

With respect to the structure of inferences, the classical syllogism comprises reasoning patterns involving two premises with a term in common (the so-called middle term) and a conclusion that involves the terms that are not shared between the premises (major term and minor term). The position of the middle term in the premises allows the definition of four different structures known as \textit{figures} (Table~\ref{tab:Aristotelianfigures}).

\begin{table}[tb]
\begin{center}
\scalebox{0.95}{
\begin{tabular}{c|c|c|c}\hline
Figure I & Figure II & Figure III & Figure IV\\ \hline\hline
$Q_{1}$ DTs are MTs & $Q_{1}$ MTs are DTs & $Q_{1}$ DTs are MTs & $Q_{1}$ MTs are DTs\\
$Q_{2}$ NTs are DTs & $Q_{2}$ NTs are DTs & $Q_{2}$ DTs are NTs & $Q_{2}$ DTs are NTs\\
-------------------- & -------------------- & -------------------- & --------------------\\
$Q$ NTs are MTs & $Q$ sNTs are MTs & $Q$ NTs are MTs & $Q$ NTs are MTs\\
\hline
\end{tabular}
}
\caption{\label{tab:Aristotelianfigures} The four figures of Aristotelian syllogistics. DT denotes middle term, MT denotes major term·and NT denotes minor term.}
\end{center}
\end{table}

Considering the four crisp quantifiers and the four figures together leads to $24$ correct inference schemes called Aristotelian \textit{moods}. All these valid cases are shown in Table~\ref{tab:Aristotelian-Moods}. They are described using the classical notation in which each mood is denoted by the symbols for the quantifiers involved, which are the major premise, the minor premise and the conclusion, respectively. For example, mood \textit{AAA} for Figure I refers to the reasoning scheme for the example in Table \ref{ex:Aristoteliansyllogism}.

\begin{table}[tb]
\begin{center}
\begin{tabular}{c|c|c|c}\hline
Figure I & Figure II & Figure III & Figure IV\\ \hline\hline
AAA & EAE & AAI & AAI\\\hline
EAE & AEE & EAO &AEE\\\hline
AII & EIO & IAI & IAI\\ \hline
EIO & AOO & AII & EAO\\ \hline
AAI & EAO & OAO & EIO\\ \hline
EAO & AEO & EIO & AEO\\ \hline
\hline
\end{tabular}
\caption{ \label{tab:Aristotelian-Moods} The four figures and 24 valid inference schemes or moods of Aristotelian syllogistics.}
\end{center}
\end{table}

Two alternative approaches have been used in the literature to endow the classical syllogism with more expressive capabilities: (i) addition of new crisp~\cite{Peterson2000} or fuzzy~\cite{Zadeh1985,Dubois1988,Novak2008} quantifiers (but preserving the number of premises) or (ii) addition of more statements to the set of premises~\cite{Sommers1982} (but only considering the four classical quantifiers).

We focus our analysis on models that follow the first alternative, in which quantifiers are extended for inclusion of fuzzy expressions, such as \textit{many, most, a few, between approximately 25\% and 30\%,\dots}, since this is the closest approach to classical Aristotelian models from the point of view of both syntax and set-based interpretation of the terms.\par

Dubois et al.~\cite{Dubois1988, Dubois1990, Dubois1993} proposed a framework in which fuzzy quantifiers are represented as intervals, as shown in the example in Table~\ref{ex:DPfuzzysyllogism}.

\begin{table}[htb]%
\centering
\begin{tabular}[c]{c}%
$[0.05, 0.1]$ people who have children are single\\
$[0.15, 0.2]$ people who have children are young\\\hline
$[0, 0.1]$ people who have children are young and single
\end{tabular}
\caption{\label{ex:DPfuzzysyllogism} Interval fuzzy syllogism.}
\end{table}

Using earlier results~\cite{Dubois1988,Dubois1990}, Dubois et al. took the first step in developing a fuzzy linguistic syllogism that is closer to the Aristotelian viewpoint than other approaches \cite{Dubois1993}.

Zadeh~\cite{Zadeh1985} defined fuzzy syllogism as ``an inference scheme in which the major premise, the minor premise and the conclusion are propositions containing fuzzy quantifiers.'' Thus, fuzzy syllogism comprises two premises and a conclusion involving quantifiers that are different to those in the LSO. A typical example of Zadeh's fuzzy syllogism is shown in Table~\ref{ex:Zadehfuzzysyllogism}. The quantifier of the conclusion, $Most^{2}:= Most \otimes Most$, is calculated from the quantifiers in the premises by applying the quantifier extension principle (QEP) ~\cite{Zadeh1983}, in this case using the fuzzy arithmetic product.\par

\begin{table}[htb]%
\centering
\begin{tabular}[c]{cc}%
$\left(P_{1}\right)  $ & Most students are young\\
$\left(P_{2}\right)  $ & Most young students are single\\\hline
$\left(C\right) $ & $Most^{2}$ students are young and single
\end{tabular}
\caption{\label{ex:Zadehfuzzysyllogism} An example of Zadeh's fuzzy syllogism.}
\end{table}

Closely related to this interpretation, Yager proposed a number of syllogistic schemes that are extensions or variants of Zadeh's framework \cite{Yager1986}. These approaches are very similar and therefore the conclusions of our analysis also directly apply to them.\par

Both approaches have been analyzed in detail in the literature \cite{Spies1989,Liu1998}. Spies considered all the models proposed by Zadeh from the point of view of their basic definitions \cite{Spies1989}. These models are syllogisms with a middle term that can be categorized into two classes:

\begin{enumerate}
\item Property inheritance (asymmetric syllogism): the link between the subject and the predicate of the conclusion is semantic. Thus, a term set $X$ and a term set $Z$ are linked via concatenation of $X$ with a term set $Y$ and $Y$ with $Z$.
\item Combination of evidence (symmetric syllogism): the link between the subject and the predicate of the conclusion is syntactic, not semantic. Thus, the links between $X$ and $Z$ and between $Y$ and $Z$ are calculated separately and they are joined in the conclusion by a logic operator (conjunction/disjunction).
\end{enumerate}

However, neither the disadvantages that QEP presents nor their compatibility with classical syllogisms are considered. Liu and Kerre analyzed the approaches of Zadeh and Dubois et al. in depth considering their multiple dimensions \cite{Liu1998}.

In this study we evaluate the capability of both fuzzy frameworks to comprise and reproduce Aristotelian moods as particular (crisp) cases. We expand the analysis of Pereira-Fari\~na et al. \cite{Pereira2010} and describe each of the syllogistic patterns in detail in terms of both their inferences schemes and the problems that arise when considering the classical inferences patterns.\par

The remainder of the paper is organized as follows. Section \ref{Sc:DbPrmodels} analyzes the reasoning patterns of Dubois et al. Section \ref{Sc:ZYmodelss} analyzes the behavior of Zadeh's patterns. In Section \ref{Sc:conclusions}, we summarize our conclusions and propose future work.

\section{Interval syllogistics: the Dubois et al. framework}
\label{Sc:DbPrmodels}
Dubois and co-workers proposed an approach to syllogistic reasoning that is based on the interpretation of linguistic quantifiers as intervals; that is, a quantifier $Q=\left[ \underline{q}, \overline{q} \right ]$, where $\underline{q}$ denotes the lower bound and $\overline{q}$ denotes the upper bound \cite{Dubois1988,Dubois1990}. Since this model focuses on managing proportional quantifiers (such as \textit{most, many, some, between 25\% and 34\%,\dots}) we have that $\underline{q}, \overline{q} \in \left[0, 1\right]$. An illustrative example of this type of syllogistic reasoning is shown in Table~\ref{ex:DPfuzzysyllogism}.

This interval approach and the Aristotelian approach have methodological differences. Aristotle manages and proves all his moods using natural logic, which involves using natural language as a logic language \cite{Sommers1982}. The approach of Dubois et al. is based on maximization and minimization of an interval (associated with the conclusion) according to other intervals given by the premises. We analyze this approach by dividing it into two parts: (i) the quantification framework; and (ii) the theory of inference.\par

\subsection{Quantification theory of Dubois et al.}

Three quantifier types are considered:

\begin{enumerate}
\item Imprecise quantifiers: quantifiers whose values are not precisely known but have precise bounds; for example, \textit{between $25\%$ and $50\%$ of students are young} and \textit{less than $70\%$ of young people are blond}. They are represented as an interval $Q=\left[ \underline{q}, \overline{q} \right]$.
\item Precise quantifiers: quantifiers whose values are precisely known and have precise bounds; for example, \textit{$10\%$ of animals are mammals} and \textit{$30\%$ of young students are tall}. In this case, the lower and upper bounds of the interval coincide ($\underline{q} = \overline{q}$).
\item Fuzzy quantifiers: quantifiers whose bounds are ill defined and are imprecise and fuzzy; for example, \textit{most Spanish cars are new} and \textit{a few elephants are pets}. Fuzzy quantifiers are represented using a fuzzy number defined through the usual four-point trapezoidal representation $Q_{i}=\left\lbrace q_{*i}, \underline{q}_{i}, \overline{q}_{i}, q^{*}_{i} \right\rbrace $, where $SUP_{Q_i}:=\left[ q_{*i},  q^{*}_{i} \right]$ represents the support of $Q_{i}$ and $KER_{Q_i}:=\left[ \underline{q}_{i}, \overline{q}_{i}\right]$ its core.
\end{enumerate}

\subsection{ Theory of inference of Dubois et al.}
\label{SSc:DbPrtheoryofinference}
The calculation procedure is based on minimization and maximization of the quantifier in the conclusion. This quantifier can be modeled as an interval or a as a trapezoidal function and is calculated by taking the quantifiers of the premises as restrictions. The main aim is to obtain the most favorable and most unfavorable proportions among the terms of the conclusion according to the proportions expressed in the premises.

From an operational point of view, researchers deal with the previously indicated pair of intervals $SUP_{Q_i}$ and $KER_{Q_i}$. Letting $A,B$ and $C$ be the labels of the term sets and $Q_{1}, Q'_{1}, Q_{2}, Q'_{2}, Q$ and $Q'$ the quantifiers, which can be precise, imprecise or fuzzy, three different reasoning schemes or patterns are proposed. We refer in more detail to Pattern I, since this has the same syntactical structure as the classical Aristotelian figures.

\subsubsection{Pattern I}

Table~\ref{tab:Dubois-PatternI} (left) shows the linguistic expression of this pattern for the three term sets \textit{A}, \textit{B} and \textit{C}, where $Q_{1}$ denotes the quantifier of the first premise and $Q'_{1}$ its converse, $Q_{2}$ denotes the quantifier of the second premise and $Q'_{2}$ its converse, and $Q$ and $Q'$ denote quantifiers for the conclusions. Table~\ref{tab:Dubois-PatternI} (right) shows an illustrative example of this pattern.

\begin{table}[htb]
\centering
\subtable{\begin{tabular} [c]{c}%
\multicolumn{1}{c}{\textbf{Linguistic expression}}\\\hline\hline
$Q_{1}\text{~As are Bs}$\\$Q_{1}^{\prime}\text{~Bs are As}$\\
$Q_{2}\text{~Bs are Cs}$\\$Q_{2}^{\prime}\text{~Cs are Bs}$\\\hline
$Q\text{~As are Cs}$\\$Q^{\prime}\text{~Cs are As}$
\end{tabular}}
\hspace{0.5cm}
\subtable{\begin{tabular} [c]{l}%
\multicolumn{1}{c}{\textbf{Example}}\\\hline\hline
$[0.85, 0.95]$ students are young\\ $[0.25, 0.35]$ young people are students\\
$[0.90, 1]$ young people are single\\ $[0.60, 0.80]$ single people are young\\\hline
\multicolumn{1}{c}{$[0.51, 1]$ students are single}
\end{tabular}}
\caption{\label{tab:Dubois-PatternI} The Pattern I syllogistic scheme of Dubois et al. Intervals in the example denote relative quantifiers.}
\end{table}

\begin{figure}
\centering
\includegraphics[width=0.50\textwidth]{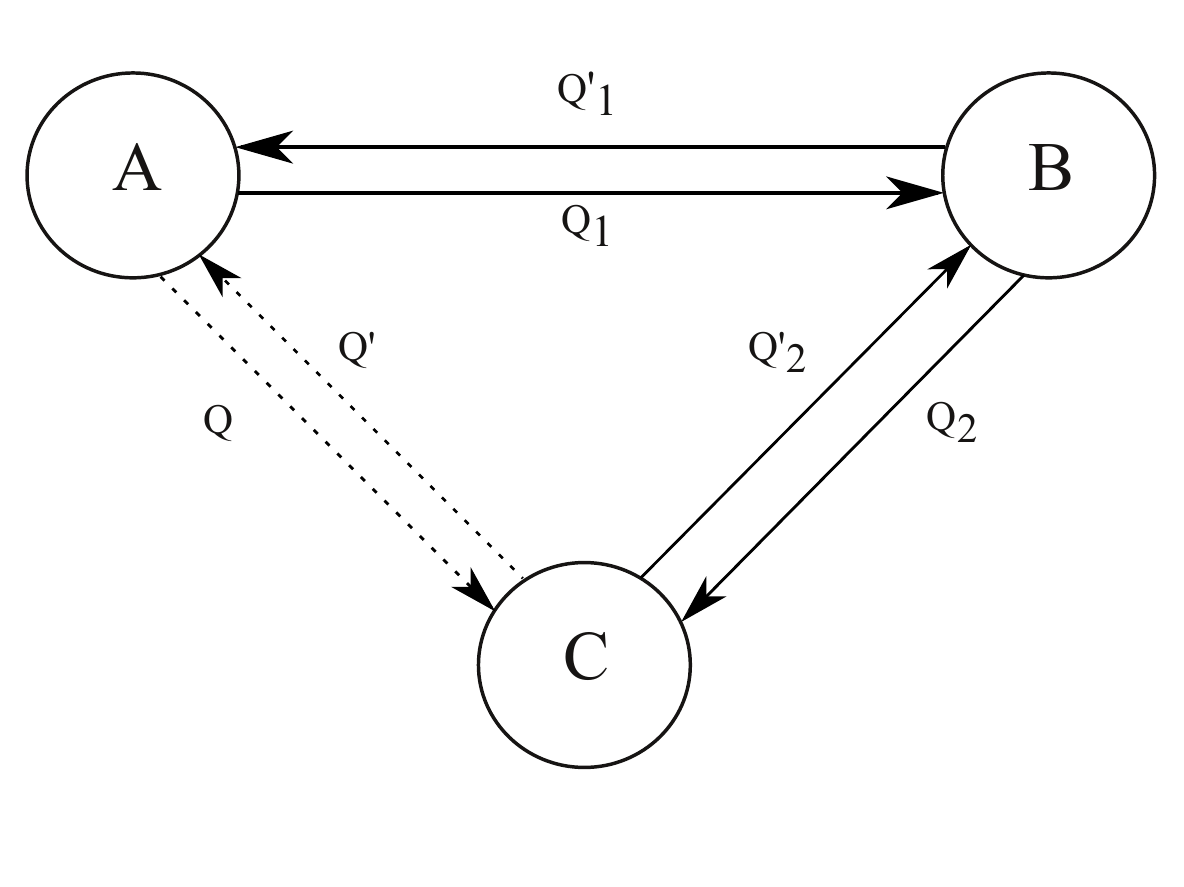}
\caption{\label{fig:Dubois-PatternI} Graphic representation of Pattern I.}
\end{figure}

Fig.~\ref{fig:Dubois-PatternI} shows a graphic representation of the Pattern I syllogistic scheme. The circles denote term sets $A, B$ and $C$; arrows with solid lines denote the links between two term sets in a premise, where $Q_{i}$ is the quantifier for that premise; arrows with discontinuous lines denote the links between two term sets in the conclusion, where $Q$ is the quantifier for the conclusion.\par

Pattern I is an asymmetric syllogism since the link between the subject ($A$) and the predicate ($C$) of the conclusion is via concatenation of $A$ with $B$ and $B$ with $C$. It is relevant to note that the links among all three predicates involved have to be fully known (i.e., the relationships between $A$ and $B$ and vice versa and between $B$ and $C$ and vice versa are needed to infer as a conclusion the relationships between $A$ and $C$ and vice versa).

For precise quantifiers, the quantifier of the conclusion $Q:=\left[\underline{q}, \overline{q} \right]$ is calculated from $Q_1:=\left[ \underline{q_1}, \overline{q_1} \right]$ and $Q_2:=\left[ \underline{q_2}, \overline{q_2} \right]$ and the corresponding $Q'_1$ and $Q'_2$ using the following expressions:\footnote{Calculation of $Q'$ is identical. For details of the proof, refer to ~\cite{Dubois1988,Dubois1990}.}

\begin{equation}
\underline{q}=q_{1} \cdotp max \left(0, 1 - \frac{1-q_{2}}{q'_{1}} \right) 
\label{Eq:Dubois-PatternI-LB}
\end{equation}

\begin{equation}
\overline{q}= min \left( 1, 1 - q_{1} + \frac{q_{1} \cdotp q_{2}}{q'_{1}}, \frac{q_{1} \cdotp q_{2}}{q'_{1} \cdotp q'_{2}}, \frac{q_{1} \cdotp q_{2}}{q'_{1} \cdotp q'_{2}} [1 - q'_{2} + q_{1}] \right) .
\label{Eq:Dubois-PatternI-UB}
\end{equation}

Management of imprecise quantifiers involves minimizing (\ref{Eq:Dubois-PatternI-LB}) and maximizing (\ref{Eq:Dubois-PatternI-UB}). The extension to fuzzy quantifiers is made by directly and independently calculating $SUP_Q$ and $KER_Q$ as a pair of imprecise quantifiers.

\subsubsection{Other patterns}
Dubois et al. proposed other reasoning schemes based on three terms \cite{Dubois1990}. Pattern II involves two linguistic expressions (Table~\ref{tab:Dubois-PatternII}) called the general version and the particular version. The main difference between them is the information available for the set of premises. The general version is applied when the information available for the terms is enough to complete the six premises; the particular version can be used when the information available allows completion of four corresponding statements. Table~\ref{tab:Dubois-example-PatternIIP} shows an illustrative example of a particular Pattern II.

\begin{table}[htb]
\centering
\subtable{\begin{tabular} [c]{l}%
\multicolumn{1}{c}{\textbf{General version}}\\\hline\hline
$Q_{1}\text{ As are Bs}$; $Q_{1}^{\prime}\text{ Bs are As}$\\
$Q_{2}\text{ Bs are Cs}$; $Q_{2}^{\prime}\text{ Cs are Bs}$\\
$Q_{3}\text{ As are Cs}$; $Q_{3}^{\prime}\text{ Cs are As}$\\\hline
$Q\text{ As and Bs are Cs}$
\end{tabular}}
\hspace{0.5cm}
\subtable{\begin{tabular} [c]{l}%
\multicolumn{1}{c}{\textbf{Particular version}}\\\hline\hline
$Q_{1}\text{ As are Bs}$; $Q_{1}^{\prime}\text{ As are Bs}$\\
$Q_{2}\text{ Bs are Cs}$\\
$Q_{3}\text{ As are Cs}$\\\hline
$Q\text{ As and Bs are Cs}$
\end{tabular}}
\caption{\label{tab:Dubois-PatternII} Linguistic schemes of Pattern II.}
\end{table}

\begin{table}[htb]%
\centering
\scalebox{0.84}{
\begin{tabular}[c]{ll}%
Between $70\%$ and $80\%$ of students are women & More than $35\%$ of women are students\\
At least $70\%$ of women are young &  \\
Between $80\%$ and $90\%$ of students are young &  \\\hline
\multicolumn{2}{c}{$Q$ of female students are young}%
\end{tabular}}
\caption{\label{tab:Dubois-example-PatternIIP} Example of a particular Pattern II.}
\end{table}

Pattern III has also two linguistic versions (Table~\ref{tab:Dubois-PatternIII}). The differences between them are the same as for Pattern II, although the number of premises changes: the general Pattern III has the same six premises but the particular one has two. Table~\ref{tab:Dubois-example-PatternIIIP} shows an example of a particular Pattern III.

\begin{table}[htb]
\centering
\subtable{\begin{tabular} [c]{l}%
\multicolumn{1}{c}{\textbf{General version}}\\\hline\hline
$Q_{1}\text{ As are Bs}$; $Q_{1}^{\prime}\text{ Bs are As}$\\
$Q_{2}\text{ Bs are Cs}$; $Q_{2}^{\prime}\text{ Cs are Bs}$\\
$Q_{3}\text{ As are Cs}$; $Q_{3}^{\prime}\text{ Cs are As}$\\\hline
$Q\text{ Cs are As and Bs}$\\
\end{tabular}}
\hspace{0.5cm}
\subtable{\begin{tabular} [c]{l}%
\multicolumn{1}{c}{\textbf{Particular version}}\\\hline\hline
$Q_{2}^{\prime}\text{ Cs are Bs}$\\
$Q_{3}^{\prime}\text{ Cs are As}$\\\hline
$Q\text{ Cs are As and Bs}$
\end{tabular}}
\caption{\label{tab:Dubois-PatternIII} Linguistic schemas of Pattern III.}
\end{table}

\begin{table}[htb]%
\centering
\scalebox{0.87}{
\begin{tabular}[c]{c}%
Between $5\%$ and $10\%$ of people who have children are single\\
Less than $5\%$ of people who have children are young\\\hline

$Q$ people who have children are young and single%
\end{tabular}}
\caption{\label{tab:Dubois-example-PatternIIIP} Example of a particular Pattern III.}
\end{table}

We can observe from the linguistic schemes for both patterns that they have a symmetric-type structure \cite{Spies1989}; that is, the core of the reasoning process is the \textit{and} logic operator. Therefore, they do not share but extend the Aristotelian inference scheme.

\subsection{Comparison of the Dubois et al. and Aristotelian syllogistics}
\label{SSc:comparisonDuboisAristotle}

Patterns II and III of Dubois et al. cannot be considered for comparison with the Aristotelian framework, since their structure is in fact a syntactical extension of the classical approach and therefore it cannot be used to reproduce classical syllogisms. Nevertheless, Pattern I can be considered for comparison, since it is an asymmetric syllogism, as are all the Aristotelian moods.

Regarding the capability of Pattern I to reproduce the four classical figures, only Figure I is compatible, as shown in Table~\ref{tab:Dubois-Figures}, when we take into account the conclusion $Q$, because of the position of the middle term. As a consequence, even though the classical Figures II, III and IV fit this approach from a syntactical point of view, they cannot be modeled within this framework.

\begin{table}
\centering
\begin{tabular}[htb]{|c|c|c|}\hline
Pattern & Major premise & Minor premise\\\hline\hline
Figure I & Subject & Predicate\\\hline
Figure II & Predicate & Predicate\\\hline
Figure III & Subject & Subject\\\hline
Figure IV & Predicate & Subject\\\hline
Pattern I & Subject & Predicate\\\hline
\end{tabular}
\caption{Position of the middle term in the Aristotelian figures and Pattern I.}
\label{tab:Dubois-Figures}
\end{table}

This limits the scope of this model and therefore comparison with the six Aristotelian moods in Figure I. To proceed with the analysis, consistent definitions for the four classical quantifiers should be provided within this framework: $A\:(all):= \left[1,1\right]$, $E\:(none):= \left[0,0\right]$, $I\:(some):= \left[\epsilon,1\right]$, $O\:(not$ $all):= \left[0, 1-\epsilon \right]$, with $\epsilon\in\left(0,1\right]$.

Using expressions (\ref{Eq:Dubois-PatternI-LB}) and (\ref{Eq:Dubois-PatternI-UB}), we obtain the results compiled in Table~\ref{tab:Dubois-patternI-MoodsFI}, showing that Pattern I is fully compatible with all the six moods.

\begin{table}
\centering
\begin{tabular}[htp] {|c|c|c|c|c|c|c|}\hline
& AAA & EAE & AII & EIO & AAI & EAO\\\hline
\textit{Pattern I} & Yes & Yes & Yes & Yes & Yes & Yes\\\hline

\end{tabular}
\caption{Behaviour of Pattern I with respect to the six moods in Figure I.}
\label{tab:Dubois-patternI-MoodsFI}
\end{table}

In conclusion, from a global point of view, the Pattern I interval approach of Dubois et al. can adequately manage the classical syllogistic moods of Aristotelian Figure I.

\section{Fuzzy syllogism: Zadeh's approach}
\label{Sc:ZYmodelss}
Table~\ref{tab:Zadeh-generalPatternFuzzySyllogism} shows the general scheme of the fuzzy syllogism proposed by Zadeh~\cite{Zadeh1985}, where $Q_{1}$, $Q_{2}$ and $Q$ are fuzzy quantifiers (\textit{most, many, some, around 25,\dots}) and $A$, $B$, $C$, $D$, $E$ and $F$ are interrelated fuzzy properties or terms.\par

\begin{table}[htb]%
\centering
\begin{tabular}[c]{c}%
\textbf{Linguistic schema}\\\hline\hline
$Q_{1}$ $As$ are $Bs$\\
$Q_{2}$ $Cs$ are $Ds$\\\hline
$Q$ $Es$ are $Fs$%
\end{tabular}
\caption{\label{tab:Zadeh-generalPatternFuzzySyllogism} Zadeh's general fuzzy syllogistic scheme.}
\end{table}

All of the reasoning patterns proposed by Zadeh are built on this general scheme by adding a number of constraints among the different fuzzy properties or terms involved in the syllogism.\par

To deepen Zadeh's approach, we divide his framework into its two fundamental components: (i) the quantification framework and (ii) the theory of inference.\par

\subsection{Zadeh's quantification framework}
\label{ssc:Zadeh-quantificationFramework}
Two linguistic quantifier types are described in this framework: absolute (\textit{around five, many, some,\dots}) and proportional (\textit{many, more than a half, most,\dots}). Thus, for instance, in ``\textit{around ten} students are blond'', \textit{around ten} is an absolute quantifier because it denotes the absolute quantity \textit{ten}; in ``\textit{most} students are blond'', \textit{most} is a proportional quantifier because it refers to a proportional quantity relative to the cardinality of the set denoted by the subject (``\textit{students}''). Some linguistic quantifiers (e.g. \textit{some}) can be either absolute or proportional, depending on the context.\par

Zadeh interpreted linguistic quantifiers as fuzzy numbers~\cite{Zadeh1983}. Therefore, absolute quantifiers are identified with absolute fuzzy numbers and proportional quantifiers with proportional fuzzy numbers.\par

The procedure proposed for combining fuzzy numbers with the corresponding fuzzy sets represented in the properties is the well-known $\Sigma Count$ scalar cardinality for fuzzy sets \cite{DeLuca1972, Zadeh1983}. It is relevant to note that the Aristotelian approach and Zadeh's approach exhibit methodological differences. As mentioned in Section \ref{Sc:DbPrmodels}, Aristotle manages and proves all his moods using natural logic~\cite{Sommers1982}, while Zadeh's inference process is based on fuzzy arithmetic, since the linguistic quantifiers are interpreted as fuzzy numbers.\footnote{Liu and Kerre discuss the most relevant problems generated by the use of QEP and fuzzy arithmetic in Zadeh's framework \cite{Liu1998}.}

\subsection{Zadeh's theory of inference}
\label{ssc:Zadeh-theoryInference}
Zadeh's approach manages the usual quantified statements ``$Q$ $As$ are $Bs$'', where $Q$ is a proportional fuzzy number (equivalent to the corresponding linguistic proportional quantifier) and $A$ and $B$ (fuzzy or crisp) sets. According to the general scheme shown in Table~\ref{tab:Zadeh-generalPatternFuzzySyllogism}, they can only be combined into inferences with two premises and a conclusion.\par

The inference process is based on QEP, which establishes that:
\begin{equation}
 \text{if } C=f(P_{1};P_{2}; \dots ;P_{n}) \text{, then } Q=\phi_{f} (Q_{1};Q_{2}; \dots ;Q_{n}) ,
\end{equation}

where $C$ is the conclusion, $P_{1};P_{2}; \dots ;P_{n}$ are the premises, $f$ is a function, $Q$ is the quantifier of the conclusion, $Q_{1};Q_{2}; \dots ;Q_{n}$ are the quantifiers of the premises and $\phi_{f}$ is an extension of $f$ obtained using the extension principle.\par

The main idea of QEP is to apply the extension principle to $f$ to obtain a fuzzy function $\phi_{f}$ that can be directly applied to the corresponding fuzzy numbers. Since fuzzy numbers are managed, the corresponding arithmetic operations must be performed using fuzzy arithmetic.\par

Now we analyze each of the syllogistic inference patterns proposed by Zadeh that emerge from the general pattern (Table~\ref{tab:Zadeh-generalPatternFuzzySyllogism}). We focus on patterns that have the same syntactical structure as the classical Aristotelian figures: multiplicative chaining and major premise reversibility chaining.

\subsubsection{Multiplicative chaining}
\label{Sssc:Zadeh-multiplicative}

Table~\ref{tab:Zadeh-chaining} (left) shows the typical linguistic expression for the multiplicative chaining (MC) reasoning pattern~\cite{Zadeh1985}. In this syllogistic pattern, $Q_{1}$ denotes the quantifier of the first premise, $Q_{2}$ the quantifier of the second premise and $Q$ the quantifier of the conclusion. Table~\ref{tab:Zadeh-chaining} (right) presents an example of use of this pattern.

\begin{figure} [htb]
\begin{center}
\includegraphics[width=0.50\textwidth]{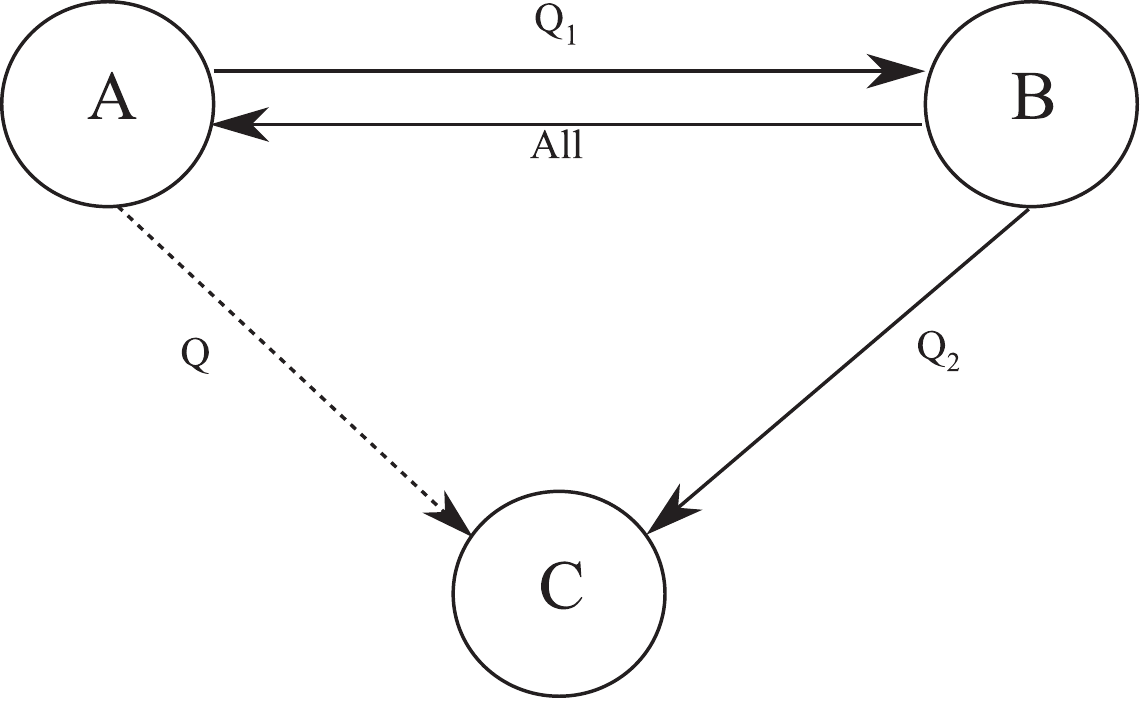}%
\caption{\label{Fg:Zadehchaining} Graphic representation of chaining syllogism.}
\end{center}
\end{figure}

Fig.~\ref{Fg:Zadehchaining} presents a graphic representation of MC syllogism. Three terms are involved that play the following roles: $A$ and $C$ constitute the conclusion, where $A$ is the subject and $C$ is the predicate, and $B$ is the ``link'' between these two terms that makes it possible to infer a coherence conclusion. Since the terms of the conclusion are ``chained'' by a middle term, this pattern is called a chaining pattern.\par

Furthermore, it is worth noting that there is a constraint between $A$ and $B$; that is, $B \subseteq A$, i.e. $\mu_{B}\left(  u_{i}\right)  \leq\mu_{A}\left(  u_{i}\right)  ,u_{i} \in U,i=1,\ldots$. This constraint can be expressed as an additional quantified statement: ``all $Bs$ as $As$''. This constraint is relevant because it allows us to know the distribution of the elements in the sets $A$ and $B$; without this information, the conclusion cannot be calculated.\par

Taking all the previous considerations into account, the procedure for calculating the conclusion is shown in Equation~(\ref{Eq:Zadeh-chaining}), where $\otimes$ denotes a fuzzy product and thus the chaining pattern is multiplicative.

\begin{equation}
Q \geq (Q_{1} \otimes Q_{2}) .
\label{Eq:Zadeh-chaining}
\end{equation}

\begin{table}[htb]
\centering
\subtable{\begin{tabular} [c]{l}%
\multicolumn{1}{c}{\textbf{Linguistic schema}}\\\hline\hline
$Q_{1}$ As are Bs (all Bs are As)\\
$Q_{2}$ Bs are Cs\\\hline
$Q$ As are Cs\\
\end{tabular}}
\hspace{0.5cm}
\subtable{\begin{tabular} [c]{l}%
\multicolumn{1}{c}{\textbf{Example}}\\\hline\hline
Most American cars are big\\
Most big cars are expensive\\\hline
\small$Most^{2}$ American cars are expensive
\end{tabular}}
\caption{\label{tab:Zadeh-chaining} Zadeh's multiplicative chaining syllogism (left) and an example (right).}
\end{table}

\subsubsection{Major premise reversibility (MPR) chaining}

Table~\ref{tab:Zadeh-MPR} (left) shows the general structure of the MPR syllogistic pattern, where $Q_{1}$ denotes the quantifier of the first premise, $Q_{2}$ the quantifier of the second premise and $Q$ the quantifier of the conclusion. Table~\ref{tab:Zadeh-MPR} (right) presents an example of the use of this pattern.

\begin{table}[htb]
\centering
\subtable{\begin{tabular} [c]{l}%
\multicolumn{1}{c}{\textbf{Linguistic scheme}}\\\hline\hline
$Q_{1}$ Bs are As\\
$Q_{2}$ Bs are Cs\\\hline
$Q$ As are Cs
\end{tabular}}
\hspace{0.5cm}
\subtable{\begin{tabular} [c]{l}%
\multicolumn{1}{c}{\textbf{Example}}\\\hline\hline
Most big cars are American\\
Most big cars are expensive\\\hline
\footnotesize$\geq0\vee\left(  2\text{ most}\ominus1\right) $ American cars are expensive
\end{tabular}}
\caption{\label{tab:Zadeh-MPR} Zadeh's major premise reversibility chaining syllogism (left) and an example of its use (right).}
\end{table}

\begin{figure}[htb]
\begin{center}
\includegraphics[width=0.50\textwidth] {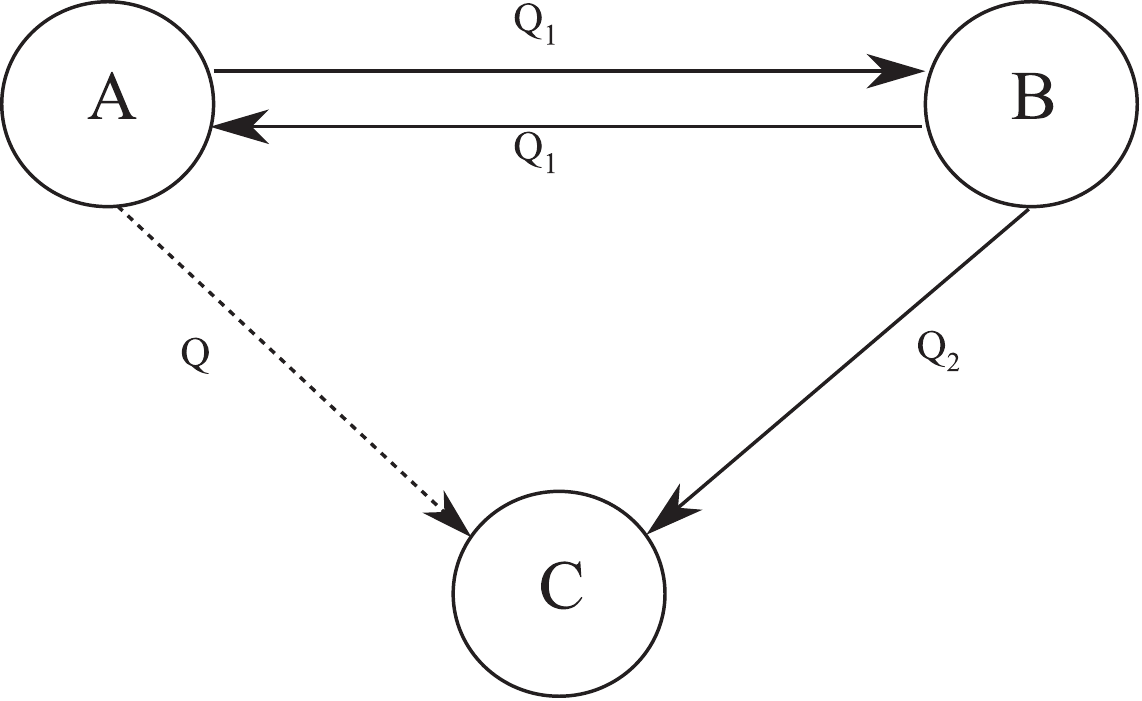}%
\caption{\label{Fg:ZadehMPRchaining} Graphic representation of MPR chaining syllogism.}
\end{center}
\end{figure}

Fig.~\ref{Fg:ZadehMPRchaining} is a graphic representation of the structure of the MPR chaining reasoning pattern~\cite{Zadeh1985}. We can consider this model as a variant of the MC pattern in Section~\ref{Sssc:Zadeh-multiplicative} where the constraint ``$B \subseteq A$'' is substituted by the reversibility of the first premise; i.e., ``$Q_{1}$ As are Bs $\leftrightarrow Q_{1}$ Bs are As''. Zadeh points out that this semantic equivalence is approximate rather than exact; the calculation of how ``approximate'' it can be remains a nontrivial open question~\cite{Zadeh1984b}.\par

Nevertheless, there is a remarkable obstacle for this constraint. For proportional quantifiers, it does not hold that a quantified sentence (e.g., ``most American cars are big'') and another sentence with the arguments interchanged (``most big cars are American'') are semantically equivalent. It is true, when talking about absolute quantifiers, that such semantic equivalence holds (e.g., ``around a hundred thousand American cars are big'' is semantically equivalent to ``around a hundred thousand big cars are American'').\par

The procedure for calculating the conclusion, shown in Equation~(\ref{eq:Zadeh-MPR}), involves fuzzy addition ($\oplus$) and subtraction ($\ominus$).

\begin{equation}
q \geq max \left(  0,q_{1}\oplus q_{2}\ominus 1 \right) .
\label{eq:Zadeh-MPR}
\end{equation}

\subsubsection{Other patterns}
Apart from the models described above, Zadeh proposed an asymmetric reasoning scheme (\textit{intersection/product}, Table~\ref{tab:Zadeh-intersection}) and two symmetric schemes (\textit{antecedent conjunction/disjunction}, Table~\ref{tab:Zadeh-antecedentC/D}; \textit{consequent conjunction/disjunction}, Table~\ref{tab:Zadeh-consequenentC/D}).

\begin{table}[htb]
\centering
\subtable{\begin{tabular} [c]{c}%
\multicolumn{1}{c}{\textbf{Linguistic schema}}\\\hline\hline
$Q_{1}$ As are Bs\\
$Q_{2}$ As and Bs are Cs\\\hline
$Q$ As are Bs and Cs
\end{tabular}}
\hspace{0.5cm}
\subtable{\begin{tabular} [c]{c}%
\multicolumn{1}{c}{\textbf{Example}}\\\hline\hline
Most students are young\\
Many young students are single\\\hline
$Q$ students are young and single
\end{tabular}}
\caption{\label{tab:Zadeh-intersection} Intersection/product syllogism.}
\end{table}

\begin{table}[htb]
\centering
\subtable{\begin{tabular} [c]{c}%
\multicolumn{1}{c}{\textbf{Linguistic scheme}}\\\hline\hline
$Q_{1}$ As are Cs\\
$Q_{2}$ Bs are Cs\\\hline
$Q$ As and/or Bs are Cs%
\end{tabular}}
\hspace{0.2cm}
\subtable{\begin{tabular} [c]{c}%
\multicolumn{1}{c}{\textbf{Example}}\\\hline\hline
Most students are young\\
Almost all single people are young\\\hline
$Q$ single people or students are young
\end{tabular}}
\caption{\label{tab:Zadeh-antecedentC/D} Antecedent conjunction/disjunction syllogism.}
\end{table}

\begin{table}[htb]
\centering
\subtable{\begin{tabular} [c]{c}%
\multicolumn{1}{c}{\textbf{Linguistic scheme}}\\\hline\hline
$Q_{1}$ As are Bs\\
$Q_{2}$ As are Cs\\\hline
$Q$ As are Bs and/or Cs
\end{tabular}}
\hspace{0.5cm}
\subtable{\begin{tabular} [c]{c}%
\multicolumn{1}{c}{\textbf{Example}}\\\hline\hline
Most students are young\\
Almost all students are single\\\hline
$Q$ students are young and single
\end{tabular}}
\caption{\label{tab:Zadeh-consequenentC/D} Consequent conjunction/disjunction syllogism.}
\end{table}

The difference between the antecedent and consequent patterns lies in the position of the logic operator in the conclusion: if it occurs for the subject, we obtain the antecedent pattern (\textit{and} for the conjunction; \textit{or} for the disjunction); if it occurs for the predicate, we obtain the consequent pattern (as in the previous pattern, the \textit{and} operator is used for the conjunction and \textit{or} for the disjunction).\par

\subsection{Comparison of Zadeh's and Aristotelian syllogistics}
\label{SSc:comparisonZadehAristotle}

We focus the analysis on the capability of Zadeh's framework to manage and reproduce Aristotelian syllogistics. First, it is worth noting that all the moods are of the property inheritance type (asymmetric syllogisms); therefore, the structure of Zadeh's symmetric syllogistic patterns is not adequate. Furthermore, the intersection/product scheme involves logic operations in the second premise and the conclusion that do not appear in the classical moods. As a consequence, none of the previously indicated patterns can be considered for this analysis.\par

Therefore, only the MC and MPR patterns can be considered for comparison with classical syllogisms. Regarding their capability to reproduce the four classical figures, only Figure I is compatible, as shown in Table~\ref{tab:Zadeh-positionMiddleTerm}, because of the position of the middle term.

\begin{table}
\centering
\begin{tabular}[htb]{|c|c|c|}
\hline
Pattern & Major premise & Minor premise\\\hline\hline
Figure I & Subject & Predicate\\\hline
Figure II & Predicate & Predicate\\\hline
Figure III & Subject & Subject\\\hline
Figure IV & Predicate & Subject\\\hline
MC & Subject & Predicate\\\hline
MPR & Subject & Predicate\\\hline
\end{tabular}
\caption{The position of the middle term in Aristotelian figures and Zadeh's MC and MPR patterns}
\label{tab:Zadeh-positionMiddleTerm}
\end{table}

As a consequence, none of the classical Figures II, III and IV can be modeled using this approach. This limits the scope of the model and therefore comparison to the six Aristotelian moods in Figure I. Zadeh's patterns are compared with these six moods in Table~\ref{tab:Zadeh-patternsMoodsFI}. Only the AII mood is compatible with the MPR scheme (i.e., one mood out of 12).

\begin{table}
\centering
\begin{tabular}[htp] {|c|c|c|c|c|c|c|}
\hline
Zadeh's patterns & AAA & EAE & AII & EIO & AAI & EAO\\\hline\hline
MC & No & No & No & No & No & No\\\hline
MPR & No & No & Yes & No & No & No\\\hline
\end{tabular}
\caption{Behavior of Zadeh's patterns with respect to Figure I.}
\label{tab:Zadeh-patternsMoodsFI}
\end{table}

The problem in the MC pattern is in the constraint $B \subseteq A$ in the first premise (minor premise in Aristotelian terms), which is much too restrictive. In this set of Aristotelian moods, the converse of the minor premise in linguistic terms is, ``some Bs are A''; that is, $A \cup B \neq \emptyset$; for instance, in the AII mood, $B \subseteq A$ cannot be inferred from the statement \textit{I: some As are Bs}. This condition is less restrictive than Zadeh's and therefore many of the possible inferences dismissed by Zadeh's pattern can be solved using the Aristotelian framework.

The MPR pattern is only valid for the AII mood since the statement and its converse in the minor premise have the same quantifier (``some''). For the other moods, those that include $A$ in the minor premise present the same problem as in the MC pattern. The solution of the EIO mood, which is compatible, presents a problem: the value obtained for $Q$ is $0$ instead of $[0,1-\epsilon)$ with $\epsilon \in (0,1]$, the usual representation for the quantifier ``not all'' in a proportional interpretation. The origin of this problem is that Zadeh's framework is not capable of managing moods involving decreasing quantifiers such as ``not all''.

In conclusion, Zadeh's approach to syllogistic reasoning is compatible with the AII mood of Aristotelian syllogism but only for the MPR scheme.

\section{Conclusions and future work}
\label{Sc:conclusions}
We have shown that the two most relevant fuzzy approaches behave very differently with regard to management of Figure I of Aristotelian syllogistics, which is one of the classical approaches to this type of reasoning. While the model of Dubois et al. is consistent with all six moods in Figure I, Zadeh's MPR scheme is only consistent for one mood out of six and his MC approach is consistent with none of the six moods. Furthermore, none of the models is compatible with the other classical Figures II, III and IV, which involve 18 moods.

Therefore, from this point of view, the proposal of Dubois et al. can be properly considered as a plausible fuzzy extension that comprises some of the classical cases as particular instances. Therefore, this proposal could be considered as a basis for defining a more general approach to syllogistic reasoning that involves, for example, the other moods currently excluded and/or other types of quantifier used in the linguistics field \cite{Diaz2003}, such as comparative (e.g., \textit{there are approximately three more tall people than blond people}) and exception (e.g., \textit{all except three students are tall} quantifiers).\par

\section*{Acknowledgment}
This work was supported in part by the Spanish Ministry of Science and Innovation (grant TIN2008-00040), the Spanish Ministry for Economy and Innovation and the European Regional Development Fund (ERDF/FEDER) (grant TIN2011-29827-C02-02) and the Spanish Ministry for Education (FPU Fellowship Program). We also wish to thank the anonymous referees for their useful and constructive comments and insightful suggestions on previous versions of this paper.

\bibliographystyle{elsarticle-num}


\end{document}